\documentclass[]{spie} 

\usepackage{amsmath,amsfonts,amssymb}
\usepackage{graphicx}
\usepackage{booktabs}
\usepackage{placeins} 
\usepackage[colorlinks=true,allcolors=blue]{hyperref}

\title{Transferable Tool--Tissue Contact Detection from Stereo Depth in Robot-Assisted Surgery}

\author[a]{Mingyeung Wu\textsuperscript{*}}
\author[a]{Zhonghao Zhang\textsuperscript{*}}
\author[b]{Hao Yang}
\author[a,b]{Alan Kuntz}
\author[b]{Jie Ying Wu\textsuperscript{\ensuremath{\dagger}}}

\affil[a]{Department of Electrical and Computer Engineering, Vanderbilt University,
Nashville, TN, United States}

\affil[b]{Department of Computer Science, Vanderbilt University,
Nashville, TN, United States}

\authorinfo{
\textsuperscript{*}These authors contributed equally to this work.\\
\textsuperscript{\ensuremath{\dagger}}Corresponding author: Jie Ying Wu,
\href{mailto:jieying.wu@vanderbilt.edu}{jieying.wu@vanderbilt.edu}.
}

\begin{document}
\maketitle

\begin{abstract}
Reliable tool--tissue contact detection can support interaction-aware control and downstream force estimation in robot-assisted surgery. Most existing methods learn a contact classifier from RGB appearance, which is hard to generalize. In this work, we use the depth image generated from a stereo pair to give more information about tool--tissue contact. For each depth frame, we localize a spatially supported minimum-distance patch around the tool boundary and reduce it to a single scalar, $-\log_{10}|d|$; this signal rises and falls in step with ground-truth contact. We formalize this observation with a fully supervised two-state hidden Markov model. We fit this model as a six-fold leave-one-session-out (LOSO) ensemble on six palpation sessions against a single silicone cup-like phantom, with the decision threshold selected from the pooled out-of-fold predictions. It is evaluated on four held-out sessions of three categories: 1. same task on same phantom; 2. same task on different phantom; 3. different task on different phantom. This model reaches held-out macro F1 $0.927$ and AUPRC $0.980$. We further compare against a reproduction of an RGB-based contact classifier from prior work. This RGB-based model achieves high performance on the first category (F1 $0.965$), but substantially lower performance on the other two, resulting in macro F1 $0.320$ across all four sessions. These results indicate that the tool--tissue distance is a strong, transferable cue for contact detection in robot-assisted surgery.

\end{abstract}


\section{INTRODUCTION}
\label{sec:introduction}
Knowing whether a surgical tool is in contact with tissue supports interaction-aware control, collision monitoring, and downstream force estimation.\cite{yang2024vision} Dedicated sensors remain difficult to sterilize, miniaturize, and integrate into existing surgical instruments.\cite{hosseinabadi2022force} Therefore, most recent work instead infers contact from the endoscopic video routinely available from the surgical system. For example, Yang et al. trained an RGB classifier to detect tool--tissue contact and used it to gate a downstream force estimate.\cite{yang2024vision} Similarly, Shibata combined kinematics-based and vision-based neural networks to detect contact for sensorless haptic feedback on the da Vinci Research Kit.\cite{shibata2025learning}
However, vision-based models are known to be difficult to generalize across domains.\cite{torralba2011unbiased} For example, an RGB-based surgical force estimation network was shown to be sensitive to shifts in camera viewpoint.\cite{chua2021toward}
We instead turn to depth: as a tool approaches tissue, the physical distance between their surfaces decreases toward zero at the moment of contact. We hypothesize that a detector built directly on this distance will generalize better.

The central claim of this paper is that the tool--tissue distance, measured between a segmented tool and its surrounding tissue surface, is a strong contact signal. We extract a spatially supported minimum-distance patch around the complete visible tool mask and reduce each frame to a scalar. To show how much of the contact signal this scalar alone carries, we deliberately turn it into a decision with a two-state Gaussian hidden Markov model with six parameters. We fit it as a six-fold leave-one-session-out (LOSO) ensemble by maximum likelihood and evaluate it on four held-out sessions of three categories: 1. same task on same phantom; 2. same task on different phantom; 3. different task on different phantom. This minimal model reaches macro F1 $0.927$ and AUPRC $0.980$ (Sec.~\ref{sec:results}), while a reproduced RGB-based contact classifier only matches this performance when the phantom and task match training. This confirms that tool--tissue distance is a strong, transferable cue for contact detection. 

\section{METHODS}
\label{sec:methods}

\subsection{Minimum-distance patch}
\label{sec:patch}

\begin{figure}[ht]
  \centering
  \includegraphics[width=\linewidth]{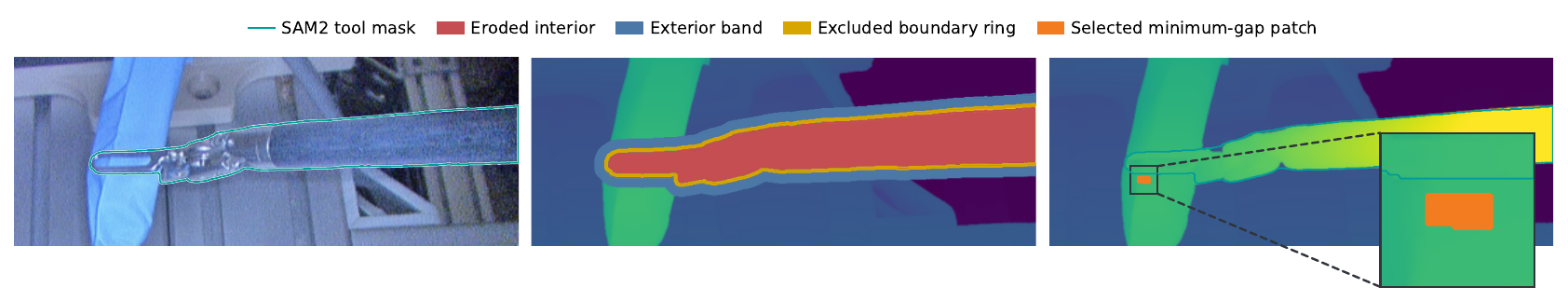}
  \caption{Patch extraction on a representative frame. Left to right: (a) SAM2 tool mask on the RGB frame; (b) eroded interior, excluded boundary ring, and exterior surface band on the depth image; (c) the selected minimum-distance patch.}
  \label{fig:pipeline}
\end{figure}

The pipeline takes a rectified stereo endoscopic frame, a per-frame Segment Anything 2 (SAM2) tool mask,\cite{ravi2025sam2} and a per-frame FoundationStereo metric-depth map\cite{wen2025foundationstereo} aligned to the same rectified frame. It outputs a single scalar: the tool--tissue distance at the \emph{minimum-distance patch} (Fig.~\ref{fig:pipeline}), defined as the region on the tool boundary closest to the surrounding tissue surface. To locate the minimum-distance patch, we (1) track the entire visible tool boundary; (2) erode the tool mask to obtain an interior; (3) build an exterior surface band whose radius scales with the visible shaft width; and (4) exclude a ring between the interior and the exterior band to reduce the effect of SAM2 mask error and FoundationStereo depth error (Fig.~\ref{fig:pipeline}). Every pixel on this exterior surface band is a candidate location for the minimum-distance patch.

Let $p$ denote such a candidate surface pixel and $q$ a neighboring pixel used in the local calculations. For each candidate $p$, we first estimate the local tool depth using a $31\times31$ window $\mathcal{N}_1(p)$ centered at $p$. Specifically, we average the depths of the valid tool pixels $q$ within this window and compare the result with the surface depth at $p$: $Z_{\mathrm{tool}}(p) = \operatorname{mean}_{q\,\in\,\mathcal{N}_1(p)\,\cap\,\mathrm{tool}} Z(q)$, and $d(p) = Z(p)-Z_{\mathrm{tool}}(p)$. The distance is computed only when $\mathcal{N}_1(p)$ contains at least $\tau_{\mathrm{tool}}=30$ valid tool pixels.

Because the distance at a single pixel can be noisy, we evaluate each candidate $p$ using a smaller $17\times17$ window $\mathcal{N}_2(p)$ centered at $p$. Each neighboring pixel $q$ in this window has its own distance value whenever $q$'s own $\mathcal{N}_1(q)$ window meets the $\tau_{\mathrm{tool}}$ requirement. We call such a $q$ \emph{valid}. We define the score of $p$ as the mean absolute distance over the valid neighboring pixels, $s(p) = \operatorname{mean}_{q\,\in\,\mathcal{N}_2(p)\,\cap\,\mathrm{valid}} \left|d(q)\right|$.

We require at least $\tau_{\mathrm{patch}}=8$ valid distances in $\mathcal{N}_2(p)$. We select the minimum-distance patch centered at $p^\ast=\arg\min_p s(p)$ (Fig.~\ref{fig:pipeline}) and report the median distance within $\mathcal{N}_2(p^\ast)$ as the tool--tissue distance for the frame. The sizes of $\mathcal{N}_1$ and $\mathcal{N}_2$ and the thresholds $\tau_{\mathrm{tool}}$ and $\tau_{\mathrm{patch}}$ are chosen jointly by a grid search over candidate window radii and pixel counts using only the six training sessions.

We hypothesize that this distance approaches zero during tool--tissue contact and increases as the tool moves freely. To compress the resulting wide range of raw distance values onto a more tractable scale, we apply a logarithmic transformation $-\log_{10}\!\left(\max\left(|d|,\epsilon\right)\right)$, where $\epsilon=10^{-4}$ prevents numerical instability near zero. After transformation, a larger value indicates a smaller distance and therefore closer tool--tissue proximity.

\subsection{A minimal supervised two-state HMM}
\label{sec:hmm}

To evaluate whether the tool--tissue distance is informative for contact detection, we model it using a two-state hidden Markov model (HMM)~\cite{rabiner1989tutorial}. The latent states correspond to non-contact and contact, each with a Gaussian emission distribution. Together with the transition probabilities, this architecture has six learned parameters. Unlike framewise thresholding, the HMM encodes short-term temporal consistency through its transition matrix, making its predictions less sensitive to isolated noisy observations. We train six HMMs under a leave-one-session-out protocol on the six training sessions. At test time, each model produces per-frame posterior contact probabilities via causal forward filtering, using only the current and past frames. These probabilities are then averaged across the six models. The decision threshold is selected from the out-of-fold training predictions and then fixed for all test sessions.

\FloatBarrier

\section{EXPERIMENTAL DESIGN}
\label{sec:experiments}

\subsection{Data collection}

Video and force data are recorded from the Patient Side Manipulator
(PSM) of a da Vinci Si surgical system (Intuitive Surgical, Inc., CA) under teleoperation by a human operator, using the open-source \texttt{dvrk\_stereo\_collection} pipeline.\cite{burkhart2026dvrkstereo}
The data collection pipeline captures left and right stereo endoscope video using Panasonic GP-U5932HT cameras (Panasonic Corporation, Osaka, Japan), digitized at $1920\times1080$ resolution in 1080i format at 30~fps. The force measurements of an ATI Gamma F/T sensor (ATI Industrial Automation, NC) are recorded together. The videos are then extracted to left/right frames and synchronized with the force sensor data using the timestamps. Frames are rectified with calibration
files, and metric depth is reconstructed from disparity produced by FoundationStereo's\cite{wen2025foundationstereo} officially released pretrained checkpoint (\texttt{23-51-11/model\_best\_bp2.pth}) on the rectified pairs. Within each session, frames are labeled as contact or non-contact by thresholding the synchronized force recording at $2\,\mathrm{N}$, except in the pulling-task session, whose frames are manually labeled as contact or non-contact instead, since the force sensor data does not provide a clear ground truth there.

\subsection{Phantoms and train/test split}

\begin{figure}[ht]
  \centering
  \begin{tabular}{cccc}
    \includegraphics[height=1.8cm]{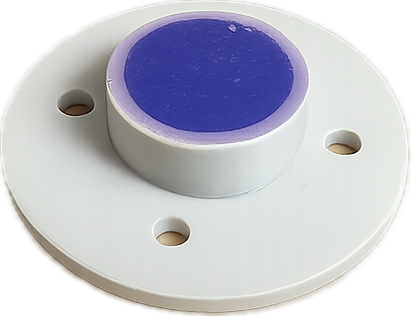} &
    \includegraphics[height=1.8cm]{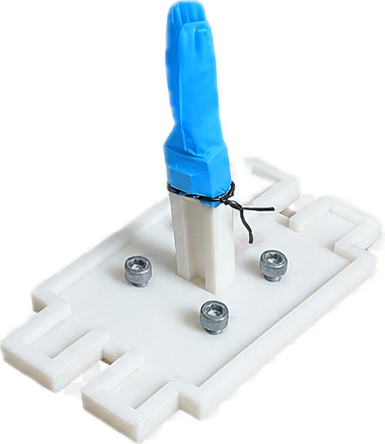} &
    \includegraphics[height=1.8cm]{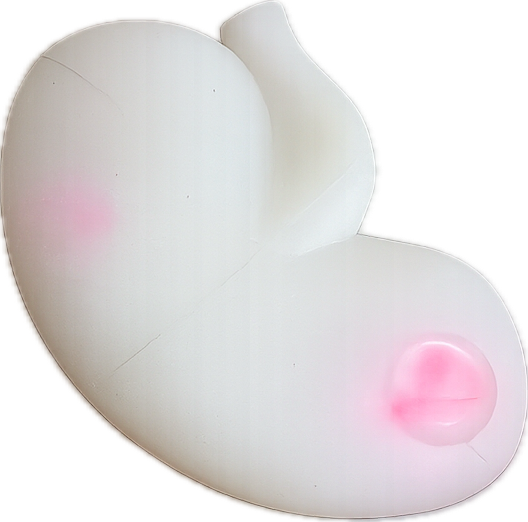} &
    \includegraphics[height=1.8cm]{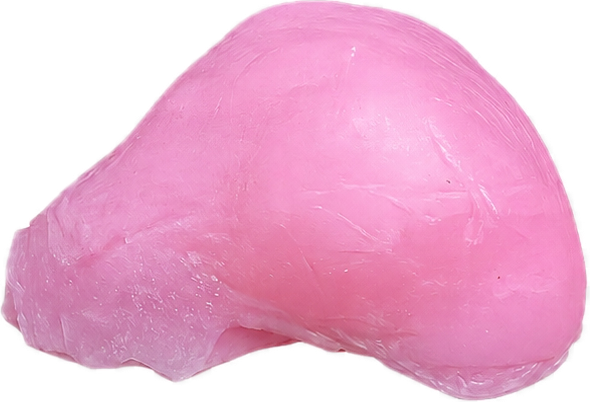} \\
    Cup & Glove & Kidney & Liver \\
  \end{tabular}
  \caption{The four phantoms used in this study.}
  \label{fig:phantoms}
\end{figure}

We collect data from four phantoms: Cup, Glove, Kidney, and Liver (Fig.~\ref{fig:phantoms}). The Cup dataset, the Kidney dataset, and the Liver dataset consist of teleoperated palpation sessions on the cup phantom, the kidney phantom, and the liver phantom. The Glove dataset contains a teleoperated pulling task on a column-like glove phantom. Six Cup sessions ($8{,}390$ frames) are used for training the HMM. Four further sessions ($2{,}417$ frames) are used for testing (one on each phantom). This split evaluates three forms of generalization: the Cup test session matches both the training phantom and task; Kidney and Liver retain the palpation task but introduce unseen phantom geometries; and Glove introduces both an unseen phantom and an unseen manipulation task.

\subsection{EfficientNet reproduction}
\label{sec:rgb_baseline}

To contextualize the HMM's performance, we reproduce the RGB-based contact classifier of Yang et al.\cite{yang2024vision} using their published implementation\cite{yang2024code}: an EfficientNetB3 model\cite{tan2019efficientnet} fine-tuned on a $234\times234$ crop centered on the tool. Three details do not carry over directly to our dataset or evaluation protocol, so we substitute the closest equivalent: (1) their DeepLabCut ``Mid 2'' keypoint crop center becomes the SAM2 tool mask's tooltip estimate; (2) their unreported learning rate becomes the standard fine-tuning default ($10^{-4}$, also used for weight decay); and (3) in place of their single split and fixed $0.5$ threshold, we follow the HMM's protocol (Sec.~\ref{sec:hmm}): 6-fold leave-one-session-out training on the Cup sessions, with pooled out-of-fold predictions used for Platt calibration and training-only F1-threshold selection, then applied to the frozen six-model ensemble on the four test sessions.

\FloatBarrier

\section{RESULTS}
\label{sec:results}

Table~\ref{tab:results} reports held-out performance on the four test sessions, ordered by increasing domain shift from training, for both the HMM (Sec.~\ref{sec:hmm}) and the EfficientNet model\cite{yang2024vision} (Sec.~\ref{sec:rgb_baseline}), under the same 6-fold LOSO ensemble protocol. The HMM maintains F1 scores of \(0.89\)--\(0.96\) and AUPRC values of \(0.97\)--\(0.99\) across the four held-out sessions. The EfficientNet model, in contrast, matches the HMM only on Cup. On Kidney and Liver its AUPRC stays high ($0.96$--$1.00$) but F1 collapses ($0.00$, $0.31$). The network still ranks contact frames correctly, but its calibrated probability rarely clears the decision threshold. On Glove, the EfficientNet model shows substantially weaker discrimination, with AUROC \(0.64\) and AUPRC \(0.75\).

\begin{table}[ht]
  \caption{Held-out performance of the HMM and the EfficientNet model\cite{yang2024vision}.}
  \label{tab:results}
  \centering
  \scriptsize
  \resizebox{\linewidth}{!}{%
  \begin{tabular}{lcccccccccccc}
    \toprule
    & \multicolumn{6}{c}{HMM} & \multicolumn{6}{c}{EfficientNet} \\
    \cmidrule(lr){2-7} \cmidrule(lr){8-13}
    Session & Accuracy & Precision & Recall & F1 & AUROC & AUPRC & Accuracy & Precision & Recall & F1 & AUROC & AUPRC \\
    \midrule
    Cup
      & 0.95 & 0.89 & 1.00 & 0.94 & 0.99 & 0.97
      & 0.97 & 0.99 & 0.94 & 0.96 & 1.00 & 1.00 \\
    Kidney
      & 0.92 & 0.84 & 1.00 & 0.91 & 0.99 & 0.98
      & 0.58 & 0.00 & 0.00 & 0.00 & 1.00 & 1.00 \\
    Liver
      & 0.90 & 0.81 & 0.99 & 0.89 & 0.99 & 0.99
      & 0.67 & 1.00 & 0.19 & 0.31 & 0.95 & 0.96 \\
    Glove
      & 0.95 & 0.93 & 1.00 & 0.96 & 0.97 & 0.98
      & 0.32 & 0.00 & 0.00 & 0.00 & 0.64 & 0.75 \\
    \textbf{Macro}
      & \textbf{0.93} & \textbf{0.87} & \textbf{1.00} & \textbf{0.93} & \textbf{0.99} & \textbf{0.98}
      & \textbf{0.63} & \textbf{0.50} & \textbf{0.28} & \textbf{0.32} & \textbf{0.90} & \textbf{0.92} \\
    \bottomrule
  \end{tabular}%
  }
\end{table}

\begin{figure}[ht]
  \centering
  \includegraphics[width=\linewidth]{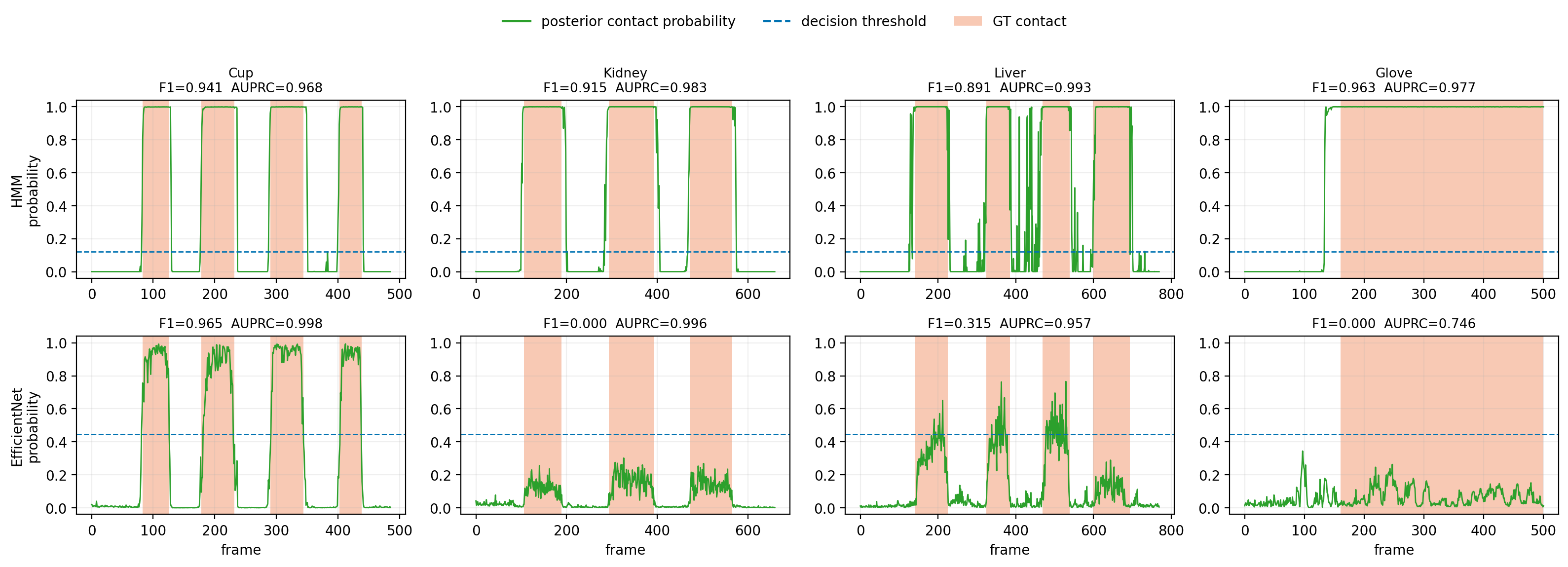}
  \caption{Predicted contact probabilities from the HMM and the EfficientNet model for the four held-out test sessions.}
  \label{fig:timelines}
\end{figure}

Fig.~\ref{fig:timelines} shows each method's predicted contact probability against ground truth: the HMM tracks contact closely in all four sessions, while the EfficientNet model rarely crosses the decision threshold on Kidney and Liver and shows substantially weaker discrimination on Glove.

\FloatBarrier

\section{DISCUSSION}
\label{sec:discussion}

The HMM generalizes across all three settings (Sec.~\ref{sec:results}), with performance staying stable as the phantom and task change. In contrast, the reproduced RGB classifier only matches this performance when the phantom and task match training. This suggests that tool--tissue distance is a simple, informative, and transferable cue for vision-based contact detection.

\subsection{Limitations}
The datasets remains limited (Sec.~\ref{sec:experiments}); since consecutive frames within a session are highly correlated, the number of sessions is the main constraint on diversity. The method depends on reliable surgical instrument segmentation, which remains challenging under variations in lighting, visual obstruction, and surgical environments.\cite{ahmed2025deep} Moreover, the quality of the reconstructed depth itself depends on stereo calibration accuracy, lighting conditions, and appropriate field of view and focus. Finally, all experiments are on silicone phantoms in relatively clean environments. Further evaluation is needed on real tissue and cluttered surgical scenes.

\subsection{Future work} The most direct extension is more diverse training data to test whether the signal continues to generalize. This includes more phantom types, more task types, and more realistic surgical scenes. With sufficient data, more expressive models (e.g., LSTMs, Transformers) could be explored with additional visual or kinematic cues. A second direction is automating the tool segmentation, which currently depends on a manual prompt for each session. Hand-eye calibration between the robot kinematics and camera could instead localize the tool automatically, removing this manual step and improving temporal consistency.

\section{CONCLUSION}
\label{sec:conclusion}

We presented a physically grounded approach to tool--tissue contact detection using the local distance between a segmented surgical tool and its surrounding tissue surface. A simple two-state HMM built solely on this signal reaches macro F1 $0.93$ and AUPRC $0.98$ across four held-out sessions spanning increasing domain shift, generalizing where a reproduced RGB-based method does not. These results highlight the utility of tool--tissue distance as a simple and generalizable cue for vision-based contact detection.

\section*{ACKNOWLEDGMENTS}
The authors thank Brendan Burkhart for developing the data collection pipeline.

\bibliographystyle{spiebib}
\bibliography{refs}

\end{document}